\documentclass[11pt,a4paper]{article}

\pdfoutput=1

\usepackage[hyperref]{acl2018}
\usepackage{times}
\usepackage{latexsym}
\usepackage{amsmath}
\usepackage{booktabs}
\usepackage{url}
\usepackage{pgfplots}
\usepackage{pgfplotstable}
\usepackage{booktabs}
\usepackage{array}
\usepackage{colortbl}
\usepackage{filecontents}
\usetikzlibrary{patterns}
\usetikzlibrary{calc,positioning}
\usepgfplotslibrary{statistics}
\usepgfplotslibrary{groupplots}
\usepackage{tikzsymbols}
\usepackage{multicol}
\usepackage{ulem}
\usepackage{microtype}
\usepackage{paralist}
  \usepackage{enumitem}
  \setlist{noitemsep}

\newcommand{\dotint}[2]{\mathrm{dot^{\prime}_{int}}(#1, #2)}

\newcommand{\quant}[1]{\mathrm{quant_{int}}(#1)}
\newcommand{\T}[1]{#1^\mathrm{T}}

\definecolor{bblue}{HTML}{4F81BD}
\definecolor{rred}{HTML}{C0504D}
\definecolor{ggreen}{HTML}{9BBB59}
\definecolor{ppurple}{HTML}{9F4C7C}
\definecolor{oorange}{HTML}{F08000}

\aclfinalcopy 

\title{Marian: Cost-effective High-Quality Neural Machine Translation in C++}

\author{Marcin Junczys-Dowmunt$^{\dagger}$ \, Kenneth Heafield$^{\ddagger}$ \\ {\bf Hieu Hoang$^{\ddagger}$ \, Roman Grundkiewicz$^{\ddagger}$ \, Anthony Aue$^{\dagger}$ } \\[3mm]
\centering
\begin{tabular}{ccc}
$^{\dagger}$Microsoft Translator && $^{\ddagger}$University of Edinburgh \\
1 Microsoft Way && 10 Crichton Street\\
Redmond, WA 98121, USA && Edinburgh, Scotland, EU\\
\end{tabular}}

\date{}

\begin{document}
\maketitle
\begin{abstract}
This paper describes the submissions of the ``Marian'' team to the WNMT 2018 shared task. We investigate combinations of teacher-student training, low-precision matrix products, auto-tuning and other methods to optimize the Transformer model on GPU and CPU.
By further integrating these methods with the new averaging attention networks, a recently introduced faster Transformer variant, we create a number of high-quality, high-performance models on the GPU and CPU, dominating the Pareto frontier for this shared task.
\end{abstract}

\section{Introduction}

This paper describes the submissions of the ``Marian'' team to the Workshop on Neural Machine Translation and Generation (WNMT 2018) shared task \cite{birch2018wnmt}.
The goal of the task is to build NMT systems on GPUs and CPUs placed on the Pareto Frontier of efficiency in accuracy.\footnote{See the shared task description: \url{https://sites.google.com/site/wnmt18/shared-task}}



Marian \cite{marian} is an efficient neural machine translation (NMT) toolkit written in pure C++ based on dynamic computation graphs.\footnote{\url{https://marian-nmt.github.io}} One of the goals of the toolkit is to provide a research tool which can be used to define state-of-the-art systems that at the same time can produce truly deployment-ready models across different devices. Ideally this should be accomplished within a single execution engine that does not require specialized, inference-only decoders. 

The CPU back-end in Marian is a very recent addition and we use the shared-task as a testing ground for various improvements. The GPU-bound computations in Marian are already highly optimized and we mostly concentrate on modeling aspects and beam-search hyper-parameters.

The weak baselines (at 16.9 BLEU on newstest2014 at least 12 BLEU points below the state-of-the-art) could promote approaches that happily sacrifice quality for speed.
We choose a quality cut-off of around 26 BLEU for the first test set (newstest2014) and do not spend much time on systems below that threshold.\footnote{We added smaller post-submission systems to demonstrate that our approach outperforms systems by other participants when we take part in the race to the quality bottom.}
This threshold was chosen based on the semi-official Sockeye \cite{DBLP:journals/corr/abs-1712-05690} baseline (27.6 BLEU on newstest2014) referenced on the shared task page.\footnote{\url{https://github.com/awslabs/sockeye/tree/wnmt18/wnmt18}} 

We believe our CPU implementation of the Transformer model \cite{NIPS2017_7181} and attention averaging networks \cite{AAN} to be the fastest reported so far. This is achieved by integer matrix multiplication with auto-tuning. We also show that these models respond very well to sequence-level knowledge-distillation methods \cite{DBLP:conf/emnlp/KimR16}. 

\section{Teacher-student training}


\subsection{State-of-the-art teacher}

Based on \newcite{DBLP:conf/emnlp/KimR16},
we first build four strong teacher models following the procedure for the Transformer-big model (model size 1024, filter size 4096, file size 813 MiB) from \newcite{NIPS2017_7181} for ensembling. We use 36,000 BPE joint subwords \cite{sennrich2016bpe} and a joint vocabulary with tied source, target, and output embeddings. One model is trained until convergence for eight days on four P40 GPUs.
See tables \ref{tab:resultsgpu} and \ref{tab:resultscpu} for BLEU scores of an overview of BLEU scores for models trained in this work.


\subsection{Interpolated sequence-level knowledge-distillation}

As described by \newcite{DBLP:conf/emnlp/KimR16}, we re-translate the full training corpus source data with the teacher ensemble as an 8-best list. Among the eight hypotheses per sentence we choose the translation with the highest sentence-level BLEU score with regard to the original target corpus. \newcite{DBLP:conf/emnlp/KimR16} refer to this method as interpolated sequence-level knowledge-distillation.  
Next, we train our student models exclusively on the newly generated and selected output.


\subsection{Decoding with small beams}
Whenever we use beam size 1, we skip softmax evaluation and simply select the output word with highest activation. The input sentences are sorted by source length, then decoded in batches of approximately equal length. We batch based on number of words. For CPU decoding we use a batch size of at least 384 words (ca. 15 sentences), for the GPU at least 8192 words (ca. 300 sentences).


\section{Student architectures}

\subsection{Transformer students}

\begin{table}[t]
\centering
\begin{tabular}{lrrrr}
\toprule
Model & Emb. & FFN & MiB \\
\midrule
Transformer-big   & 1024 & 4096 & 813 \\
Transformer-base  & 512 & 2048  & 238 \\
Transformer-small & 256 & 2048  & 101 \\
Transformer-tiny-256*  & 256 & 1536  & 84\\
Transformer-tiny-192*  & 192 & 1536  & 60\\
\bottomrule
\end{tabular}
\caption{Transformer students dimensions. Post-submission models marked with *.}\label{trans.dim}
\end{table}

For our Transformer student models we follow the Transformer-big and Transformer-base configurations from \newcite{NIPS2017_7181}. Additionally we investigate a Transformer-small and post-submission two Transformer-tiny variants on the CPU. We also use six blocks of self-attention, source-attention, and FFN layers with varying embedding (model) and FNN sizes, see Table~\ref{trans.dim}.

Transformer-big is initialized with one of the original teachers and fine-tuned on the teacher-generated data until development set BLEU stops improving for beam-size 1. The remaining student models are trained from scratch on teacher-generated data until development set BLEU stalls for 20 validation steps when using beam-size~1.

\subsection{Averaging attention networks}
Very recently, \newcite{AAN} suggested averaging attention networks (AAN), a modification of the original Transformer model that addresses a decode-time inefficiency, apparently without loss of quality.
During translation, the self-attention layers in the Transformer decoder look back at their entire history, introducing quadratic complexity with respect to output length.
\newcite{AAN} replace the decoder self-attention layer with a cumulative uniform averaging operation across the previous layer. During decoding, this operation can be computed based on the single last step. Decoding is then linear with respect to output length. \newcite{AAN} also add a feed-forward network and a gate to the block. We choose a smaller FFN size than \newcite{AAN} (corresponding to embeddings size instead of FFN size in table \ref{trans.dim}) and experiment with removing the FFN and gate.

\subsection{RNN-based students}

Our focus lies on efficient CPU-bound Transformer implementations. However, Marian and its predecessor Amun \cite{amun} were first implemented as fast GPU-bound implementations of Nematus-style \cite{sennrich2017nematus} RNN-based translation models. We use these models to cover the lower end of the quality spectrum in the task. We train a standard shallow GRU model (RNN-Nematus, embedding size 512, state size 1024), a small version (RNN-small, embedding size 256, state size 512) and a deep version with 4 stacked GRU blocks in the encoder and 8 stacked GRU blocks in the decoder (RNN-deep, embedding size 512, states size 1024). This model corresponds to the University of Edinburgh submission to WMT 2017
\cite{sennrich-EtAl:2017:WMT}.

\section{Optimizing for the CPU}
Most of our effort was concentrated on improving CPU computation in Marian. Apart from improvements from code profiling and bottleneck identification,
we worked towards integrating integer-based matrix products into Marian's computation graphs.

\subsection{Shortlist}

A simple way to improve CPU-bound NMT efficiency is to restrict the final output matrix multiplication to a small subset of translation candidates. We use a shortlist created with fastalign \cite{DBLP:conf/naacl/DyerCS13}. For every mini-batch we restrict the output vocabulary to the union of the 100 most frequent target words and the 100 most probable translations for every source word in a batch. All CPU results are computed with a shortlist.


\subsection{Quantization and integer products}
Previously, Marian tensors would only work with 32-bit floating point numbers. We now support tensors with underlying types corresponding to the standard numerical types in C++. We focus on integer tensors.


Some of our submissions replaced 32-bit floating-point matrix multiplication with 16-bit or 8-bit signed integers.  For 16-bit integers, we follow \newcite{sharp} in simply multiplying parameters and inputs by $2^{10}$ before rounding to signed integers.  This does not use the full range of values of a 16-bit integer so as to prevent overflow when accumulating 32-bit sums; there is no AVX512F instruction for 32-bit add with saturation.

For 8-bit integers, we swept quantization multipliers and found that 29 was optimal, but quality was still poor.  Instead, we retrained the model with matrix product inputs (activations and parameters but not outputs) clipped to a range.  We tried $[-3,3]$, $[-2,2]$, and $[-1, 1]$ then settled on $[-2,2]$ because it had slightly better BLEU.\footnote{This might however have been an artifact of the posterior clipping process rather than an effect of quantization.}  Values were then scaled linearly to $[-127, 127]$ and rounded to integers.  We accumulated in 16-bit integers with saturation because this was faster, observing a 0.05\% BLEU drop relative to 32-bit accumulation.


The test CPU is a Xeon Platinum 8175M with support for AVX512.  We used these instructions to implement matrix multiplication over 32 16-bit integers or 64 8-bit integers at a time.\footnote{The only packed 8-bit multiplication instruction is \texttt{vpmaddubsw}, which requires AVX512BW.  Interestingly, Amazon's hypervisor hides support for AVX512BW from CPUID but the instruction works as expected so we used it.}

\subsection{Memoization}
To ensure contiguous memory access, the integer matrix product $\dotint{A}{B}$ calculates $A\T{B}$ instead of $AB$. It also expects its inputs $A$ and $B$ to be correctly quantized integer tensors. Therefore, we have to compute $\dotint{\quant{A}}{\quant{\T{B}}}$ to use the quantized integer product as a replacement for the floating point matrix product.

In most cases, $B$ is a parameter, while $A$ contains activations. Repeating the quantization and transposition operations for every decoder parameter at every step would incur a significant performance penalty. To counter this, we introduce memoization into Marian's computation graphs. Memoization caches the values of constant nodes that will not change during the lifetime of the graph.

During inference, parameter nodes are constant.
Apart from that any node with only constant children is constant and can be memoized. In our example, $B$ is constant as a parameter, $\T{B}$ is constant because its only child is constant, so is $\quant{\T{B}}$. $\dotint{\quant{A}}{\quant{\T{B}}}$ itself is not constant, as the activations $A$ can change. Values for constant nodes are calculated only once during the first forward step in which they appear; subsequent calls will use cached versions. 


\subsection{Auto-tuning}

\begin{table}[t]
\centering
\begin{tabular}{lrrr} \toprule
Model & 1s & 384w & BLEU \\ \midrule
Transf.-base-AAN & 1018.8 & 397.5 & 27.5 \\
+shortlist & 758.1  & 293.7 & 27.5\\
+int16          & 2703.2 & 491.4 & 27.5\\
+memoization    & 572.9  & 294.3 & 27.5\\
+auto-tuning    & 574.8  & 273.2 & 27.5 \\
\midrule
Transformer-big & 4797.0 & 1537.8 & 28.1 \\
+clip=2 (+mem.) & 5006.9 & 1737.1 & 27.7 \\ 
+int8 (+mem.)   & 1772.6 & 1169.9 & 27.5 \\
\bottomrule
\end{tabular}
\caption{Time to translate newstest2014 with batch-size equal to 1 sentence (1s) and around 384 words (384w) using integer multiplication variants vs 32-bit float matrix multiplication. \label{tab.opt}}
\end{table}

At this point, the float32 (Intel's MKL) product and our int16 matrix product can be used interchangeably for small and mid-sized models (we see overflow for the large Transformer model). While trying to choose one implementation, we noticed that both algorithms will outperform the respective other in different contexts. In the face of many different matrix sizes and access patterns it is difficult to determine reliable performance profiles. Instead, we implemented an auto-tuner.

We hash tensor shapes and algorithm IDs and annotate each node in an alternative subgraph with a timer. We collect the total execution time across 100 traversals of each alternate subgraph. Once this limit has been reached, usually within a few sentences, the auto-tuner stops measurements and selects the fastest alternative for all subsequent calls.

\subsection{Optimization results}
Table \ref{tab.opt} illustrates the effects of the optimizations introduced in this section for sentence-by-sentence and batched translation. Adding a shortlist improves translation speed significantly. Enabling int16 multiplication without memoization hurts performance; with memoization we see improvements for single-sentence translation and similar performance to MKL for batched translation. With auto-tuning, single-sentence translation achieves the same performance as before and batched translation improves. In both cases the auto-tuning algorithm was able to choose a good solution. In the single-sentence case we would always use the int16 product. In the batched case a mix performs better than a hard choice.

We also see respectable improvements for the Transformer-big model with int8 multiplication. Most of the loss in BLEU is due to the fine-tuning process with clipping during training. 


\section{Results and cost-effective decoding}

\begin{table*}[p]
\centering\renewcommand{\arraystretch}{0.95}
\begin{tabular}{cp{7cm}rrr}
\toprule
No & Model & MiB & Time & BLEU \\
\midrule
(1) & Baseline GPU & -- & 51.6 & 16.8\\
(2) & Sockeye GPU (Transformer-base b=5) & -- & 231.9 & 27.6 \\
\midrule
(3) & Teacher - Transformer-big b=8 & 813 & 109.7 & 28.1 \\
(4) & Teacher - Transformer-big$\times 4$ b=8 & 3252 & 410.8 & 29.0 \\
\midrule
(5) & Transformer-big b=4 & 813 & 52.0 & 28.4 \\ 
(6) & \bf Transformer-big b=2 & \bf 813 & \bf 31.9 & \bf 28.4 \\ 
(7) & Transformer-big & 813 & 19.9 & 28.2 \\  \midrule 
(8) & Transformer-base b=4 & 238 & 40.5 & 27.8 \\ 
(9) & Transformer-base b=2 & 238 & 22.9 &  27.8 \\ 
(10) & Transformer-base & 238 & 12.8 & 27.6 \\ \midrule 
(11) & Transformer-base-AAN b=4 & 220 & 15.9 & 27.7 \\
(12) & \bf Transformer-base-AAN b=2 &\bf 220 &\bf 8.9 &\bf 27.7 \\
(13) & Transformer-base-AAN & 220 & 7.2 & 27.6 \\ \midrule
(14) & Transformer-small & 101 & 10.8 & 26.4 \\ \midrule
(15) & Transformer-small-AAN & 100 & 5.9 & 25.8 \\
(16) & \bf Transformer-small-AAN -ffn & \bf 98 & \bf 5.7 & \bf 26.2 \\
(17) & Transformer-small-AAN -ffn -gate & 95 & 5.6 & 25.8 \\ \midrule
(18) & \bf RNN-small-Amun & \bf 88 & \bf 1.6 & \bf 24.1 \\ 
(19) & RNN-Nematus-Amun & 199 & 2.2 & 24.8 \\ \midrule 
(20) & RNN-small& 88 & 1.8 & 24.1 \\ 
(21) & RNN-Nematus & 199 & 2.5 & 24.8 \\ 
(22) & RNN-Deep & 323 & 2.9 & 25.7 \\ 
\bottomrule
\end{tabular}
\caption{Results on newstest2014 - GPU systems. Submitted systems in bold. All student systems have been used with beam-size 1 unless stated differently (b=$n$).
}\label{tab:resultsgpu}
\end{table*}

\begin{table*}[p]
\centering\renewcommand{\arraystretch}{0.95}
\begin{tabular}{cp{7cm}rrr}
\toprule
No & Model & MiB & Time & BLEU \\
\midrule
(1) & Baseline CPU & -- & 4492.2 & 16.8 \\
(2) & Sockeye CPU (Transformer-base b=5) & -- & 1168.6 & 27.4 \\ \midrule
(7) & \bf Transformer-big & \bf 813 & \bf 1537.8 & \bf 28.1 \\
(7i) & \bf Transformer-big-int8 & \bf 813 & \bf 1169.9 & \bf 27.5 \\ \midrule
(10) & Transformer-base & 238 & 393.1 & 27.4 \\
(10i) & Transformer-base-int16 & 238 & 400.2 & 27.4 \\ \midrule
(13) & Transformer-base-AAN & 220 & 288.7 &  27.5 \\
(13i) & \bf Transformer-base-AAN-int16 & \bf 220 & \bf 273.2 & \bf 27.5 \\ \midrule
(14) & Transformer-small & 101 & 134.1 & 26.5 \\
(14i) & Transformer-small-int16 & 101 & 133.2 & 26.5 \\  \midrule
(15i) & Transformer-small-AAN-int16 & 100 & 108.8 & 25.8 \\
(16i) & Transformer-small-AAN-int16 -ffn  & 98 & 108.3 & 26.2 \\
(17) & Transformer-small-AAN -ffn -gate & 95 & 100.6 & 26.0 \\
(17i) & \bf Transformer-small-AAN-int16 -ffn -gate & \bf 95 & \bf 94.1 & \bf 26.0 \\ \midrule
(23i) & Transformer-tiny-256-AAN-int16 -ffn -gate* & 84 & 79.7 & 25.3 \\
(24i) & Transformer-tiny-192-AAN-int16 -ffn -gate* & 60 & 61.1 & 24.4 \\
\bottomrule
\end{tabular}
\caption{Results on newstest2014 - CPU systems. Submitted systems in bold. Post-submission systems marked with *.
All student systems have been used with beam-size 1 unless stated differently (b=$n$).
}\label{tab:resultscpu}
\end{table*}

In tables \ref{tab:resultsgpu} and \ref{tab:resultscpu}, we summarize our experiments with GPU and CPU models. Bold rows contain results for our task submissions. We report model sizes in MiB, translation time without initialization and BLEU scores for newstest2014. Time has been measured on AWS p3.x2large instances (NVidia V100) and AWS m5.large instances, the official evaluation platforms of the shared task.

All our student models outperform the baselines in terms of translation speed and quality, but as stated before, we are mostly interested in models above a 26 BLEU threshold. It seems that the new AAN architecture is a promising modification of the Transformer with minimal or no quality loss in comparison to its standard equivalent. We also see that teacher-student methods can be successfully used to create high-performance and high-quality Transformer systems with greedy decoding.

\begin{figure*}[p]
\centering
\begin{tikzpicture}

\begin{axis}[
scale=.95,
xmode=log,
xmin=0.08, xmax=125,
xtick={0.1, 1, 10, 100},
log ticks with fixed point,
every non boxed x axis/.style={},
scale only axis,
ymajorgrids, yminorgrids=false,
xmajorgrids, xminorgrids,
x axis line style=-,
group style={
group name=break2,
group size=1 by 2,
xticklabels at=edge bottom,
vertical sep=0pt
},
minor grid style={gray!30, ultra thin},
y tick label style={/pgf/number format/fixed,
/pgf/number format/fixed zerofill,
/pgf/number format/precision=1},
xtick distance=5,
visualization depends on={value \thisrow{sysid} \as \thesystem},
nodes near coords=\thesystem,
every node near coord/.append style={anchor=north, font=\scriptsize, xshift=0pt},
legend columns=2,
legend style={at={(0.5,1.03)},anchor=north,font=\small, /tikz/column 2/.style={column sep=5pt},},
width=14cm,
height=10cm,
ymin=16.0, ymax=30.5,
restrict y to domain=16.5:29.5,
ylabel={BLEU}, xlabel={Million translated source tokens per USD (log scale)},
]

\addplot [only marks, bblue, mark=x, very thick, mark size=3]
table[x expr={62954 / \thisrow{time} * 3600/3.259 / 10^6}, y=bleu] {gpu.systems.marian};

\addplot [only marks, rred, mark=x, very thick, mark size=3]
table[x expr={62954 / \thisrow{timed} * 3600/0.102 / 10^6}, y=bleu] {cpu.systems.marian};

\addplot [only marks, bblue, mark=o, very thick, mark size=3]
table[x expr={62954 / \thisrow{time} * 3600/3.259 / 10^6}, y=bleu] {gpu.systems.other};

\addplot [only marks, rred, mark=o, very thick, mark size=3]
table[x expr={62954 / \thisrow{time} * 3600/0.102 / 10^6}, y=bleu] {cpu.systems.other};

\addplot [only marks, black, mark=square, mark size=3, nodes near coords={}]
table[x expr={62954 / \thisrow{time} * 3600/3.259 / 10^6}, y=bleu] {gpu.systems.marian.submitted};
\node at (axis cs:85,29.8) {\Cooley[1.5]};

\addplot [only marks, black, mark=square, mark size=3, nodes near coords={}]
table[x expr={62954 / \thisrow{timed} * 3600/0.102 / 10^6}, y=bleu] {cpu.systems.marian.submitted};

\addplot [only marks, rred, mark=o, very thick, mark size=3, nodes near coords={}]
table[x expr={62954 / \thisrow{time} * 3600/0.102 / 10^6}, y=bleu] {cpu.systems.other2};

\addplot [only marks, bblue, mark=o, very thick, mark size=3, nodes near coords={}]
table[x expr={62954 / \thisrow{time} * 3600/3.259 / 10^6}, y=bleu] {gpu.systems.other2};

\node at (axis cs:85,29.8) {\Cooley[1.5]};

\legend{Marian GPU, Marian CPU, Others GPU, Others CPU, Submissions}

\end{axis}
\end{tikzpicture}
\caption{Cost-effectiveness (logarithmic scale) vs BLEU for all systems and baselines. }\label{effect.log}

\end{figure*}

\begin{figure*}[p]
\centering
\begin{tikzpicture}

\begin{groupplot}[
scale=0.95,
every non boxed x axis/.style={},
scale only axis,
width=14cm,
ymajorgrids, yminorgrids=false,
xmajorgrids, 
minor x tick num=4,
xmin=-.5, xmax=25.5,
x axis line style=-,
group style={
group name=break2,
group size=1 by 2,
xticklabels at=edge bottom,
vertical sep=0pt
},
minor grid style={gray!30, ultra thin},
y tick label style={/pgf/number format/fixed,
/pgf/number format/fixed zerofill,
/pgf/number format/precision=1},
xtick distance=5,
ylabel={BLEU}, xlabel={},
visualization depends on={value \thisrow{sysid} \as \thesystem},
nodes near coords=\thesystem,
every node near coord/.append style={anchor=north, font=\scriptsize, xshift=0pt},
legend columns=2,
legend style={at={(0.5,1.03)},anchor=north,font=\small, /tikz/column 2/.style={column sep=5pt},},
every axis y label/.append style={at={(-0.02,0.3)}}
]
\nextgroupplot[
height=8cm,
ymin=25.5, ymax=29.5,
restrict y to domain=25.5:29.5,
axis x line=top,
axis y discontinuity=crunch,
ytick={26,26.5,27,27.5,28,28.5,29},
yticklabels={26.0,26.5,27.0,27.5,28.0,28.5,29.0},
]

\addplot [only marks, bblue, mark=x, very thick, mark size=3]
table[x expr={62954 / \thisrow{time} * 3600/3.259 / 10^6}, y=bleu] {gpu.systems.marian};

\addplot [only marks, rred, mark=x, very thick, mark size=3]
table[x expr={62954 / \thisrow{timed} * 3600/0.102 / 10^6}, y=bleu] {cpu.systems.marian};



\addplot [only marks, bblue, mark=o, very thick, mark size=3]
table[x expr={62954 / \thisrow{time} * 3600/3.259 / 10^6}, y=bleu] {gpu.systems.other};

\addplot [only marks, rred, mark=o, very thick, mark size=3]
table[x expr={62954 / \thisrow{time} * 3600/0.102 / 10^6}, y=bleu] {cpu.systems.other};

\addplot [only marks, black, mark=square, mark size=3, nodes near coords={}]
table[x expr={62954 / \thisrow{time} * 3600/3.259 / 10^6}, y=bleu] {gpu.systems.marian.submitted};

\addplot [only marks, black, mark=square, mark size=3, nodes near coords={}]
table[x expr={62954 / \thisrow{timed} * 3600/0.102 / 10^6}, y=bleu] {cpu.systems.marian.submitted};

\addplot [only marks, rred, mark=o, very thick, mark size=3, nodes near coords={}]
table[x expr={62954 / \thisrow{time} * 3600/0.102 / 10^6}, y=bleu]{cpu.systems.other2};

\addplot [only marks, bblue, mark=o, very thick, mark size=3, nodes near coords={}]
table[x expr={62954 / \thisrow{time} * 3600/3.259 / 10^6}, y=bleu] {gpu.systems.other2};

\node at (axis cs:24,29.2) {\Cooley[1.5]};

\legend{Marian GPU, Marian CPU, Others GPU, Others CPU, Submissions}

\nextgroupplot[
height=2cm,
ymin=16.25, ymax=17.25,
ylabel={}, xlabel={Million translated source tokens per USD},
axis x line=bottom,
ytick={16.5,17,17.5},
yticklabels={16.5, 17.0,},
]

\addplot [only marks, bblue, mark=x, very thick,, mark size=3]
table[x expr={62954 / \thisrow{time} * 3600/3.259 / 10^6}, y=bleu] {gpu.systems.marian};

\addplot [only marks, bblue, mark=o, very thick,, mark size=3]
table[x expr={62954 / \thisrow{time} * 3600/3.259 / 10^6}, y=bleu] {gpu.systems.other};

\addplot [only marks, rred, mark=o, very thick,, mark size=3] table[x expr={62954 / \thisrow{time} * 3600/0.102 / 10^6}, y=bleu] {cpu.systems.other};

\addplot [only marks, rred, mark=x, very thick,, mark size=3]
table[x expr={62954 / \thisrow{time} * 3600/0.102 / 10^6}, y=bleu] {cpu.systems.marian};

\end{groupplot}
\end{tikzpicture}
\caption{Cost-effectiveness (linear scale) vs BLEU for systems around and above 26 BLEU and baselines.}\label{effect.lin}

\end{figure*}

We  compare our systems on a common cost-effectiveness scale expressed as the number of source tokens translated per US Dollar $\left[\mathrm{\frac{w}{USD}}\right]$. Given the hourly price for a dedicated AWS GPU (p3.x2large, 3.259 USD/h) or CPU (m5.large, 0.102 USD/h) instance\footnote{The same instance types were used for the shared task.} and the time to translate newstest2014 consisting of 62,954 source tokens with a chosen model and instance, we calculate:
$$ \frac{\mathrm{62,954\;[w]}}{\mathrm{Translation\;time\;[s]}} \cdot \frac{\mathrm{3,600\;\left[s/h\right]}}{\mathrm{Instance\; price\;\left[\$/h\right]}}. $$

This representation has multiple advantages:
\begin{itemize}
\item Systems deployed on different hardware can be compared directly;
\item The linear mappings into the common space are scale-preserving and correctly represent relative speed differences between systems on the same hardware;
\item We can relate three important categories --- speed, quality, and cost --- to each other in a single visualization.
\end{itemize}

Figures \ref{effect.log} and \ref{effect.lin} illustrate cost-effectiveness of our models, the baselines and submissions by other participants versus translation quality on newstest2014. Figure \ref{effect.log} contains all models with a cost-effectiveness log-scale. This reflects a trend that speed gains are exponential in quality loss. Based on Figure~\ref{effect.log}, it seems that our models dominate the Pareto-frontier for high-quality models for CPU and GPU models compared to the baselines and other participants.

We added post-submission systems (23i) and (24i) on the CPU to demonstrate that we can outperform the results of other participants for speed and quality when lowering our quality threshold.

In Figure~\ref{effect.lin} with a linear cost-effectiveness scale, we emphasize models around and above the quality threshold of 26 BLEU which were our main focus in this work. It is interesting to see that similar Marian models have surprisingly similar cost-effectiveness across different hardware types.


\section{Conclusions}

We demonstrated that Marian can serve as an integrated research and deployment platform with highly efficient decoding algorithms on the GPU and CPU.
Transformer architectures can be efficiently trained in teacher-student settings and then used with small beams or with greedy decoding. To our knowledge, this is also the first work to integrate Transformer architectures with low-precision matrix multiplication. By combining these methods with the new averaging attention networks, we created a number of high-quality, high-performance models on the GPU and CPU, dominating the Pareto frontier for this shared task. 



\section*{Acknowledgments}
We thank Jon Clark who suggested the graphing in USD.  Partly supported by a Mozilla Research Grant.

\bibliography{acl2018}
\bibliographystyle{acl_natbib}

\end{document}